\documentclass[journal]{IEEEtran}
\usepackage{graphicx}
\usepackage{subfig}
\usepackage{graphicx}
\usepackage{booktabs}
\usepackage{comment}
\usepackage{color}
\graphicspath{ {images/} }
\usepackage{subfig}
\begin{document}
\title{SkinNet: A Deep Learning Framework for Skin Lesion Segmentation}

\author{Sulaiman Vesal,
	Nishant Ravikumar
	and~Andreas Maier% <-this % stops a space
	% <-this % stops a space
	\thanks{S. Vesal, N.Ravikumar, and A. Maier  are with the Pattern Recognition Lab, Friedrich-Alexander-University Erlangen-Nuremberg, Erlangen, Germany (telephone: +49 152 10228542, e-mail: \{sulaiman.vesal, nishant.kumar, andreas.maier\}@fau.de).}%
}
\maketitle

\begin{abstract}
There has been a steady increase in the incidence of skin cancer worldwide, with a high rate of mortality. Early detection and segmentation of skin lesions is crucial for timely diagnosis and treatment, necessary to improve the survival rate of patients. However, skin lesion segmentation is a challenging task due to the low contrast of lesions and their high similarity in terms of appearance, to healthy tissue. This underlines the need for an accurate and automatic approach for skin lesion segmentation. To tackle this issue, we propose a convolutional neural network (CNN) called SkinNet.
The proposed CNN is a modified version of U-Net. We compared the performance of our approach with other state-of-the-art techniques, using the ISBI 2017 challenge dataset. Our approach outperformed the others in terms of the Dice coefficient, Jaccard index and sensitivity, evaluated on the held-out challenge test data set, across 5-fold cross validation experiments. SkinNet achieved an average value of 85.10, 76.67 and 93\%, for the DC, JI and SE, respectively.

\end{abstract}

%\begin{IEEEkeywords}
%IEEEtran, journal, \LaTeX, paper, template.
%\end{IEEEkeywords}

\section{Introduction}

\IEEEPARstart{O}{ver} 5 million new cases of skin cancer are diagnosed in the United States alone, each year. Melanoma is the advanced form of skin cancer and the global incidence of melanoma was estimated to be over 350,000 cases in 2015, with almost 60,000 deaths. Although the mortality rate is significant, early detection improves the survival rate to over 95\%\cite{s18020556}\cite{ISBI}. Computer-aided-diagnostic systems that enable automatic and accurate skin lesion detection, segmentation and classification are thus essential. Recently, many studies attempted to address this challenge, for example: \cite{s18020556} proposed a deep learning framework consisting of two fully convolutional residual networks, to simultaneously segment and classify skin lesions; \cite{Ma} introduced a deformable model using a newly defined speed function and stopping criterion for skin lesion segmentation; and \cite{Yu} used a deep learning approach called fully convolutional residual network (FCRN) with more than 50 layers for both segmentation and classification. 
In this study we propose a CNN-based skin lesion segmentation framework, called SkinNet. The proposed CNN architecture is a modified version of the U-Net \cite{Unett}. The latter has demonstrated state-of-the-art performance in various medical image segmentation tasks in recent years. The U-net architecture basically consists of a contracting path (encoder), which downsamples an image into a set of high-level features, followed by a symmetric expanding path (decoder), which uses the feature information to build a pixel-wise segmentation mask. SkinNet employs dilated convolutions \cite{Dilated} in the lowest layer of the encoder-branch in the U-Net, to provide a more global context for the features extracted from the image. Additionally, we replace the conventional convolution layers in both the encoder and decoder branches of U-Net, with dense convolution blocks, to better incorporate multi-scale image information. Section \ref{sec:methods} describes the proposed method, dataset and experimental setup and the results are summarized in Section \ref{sec:conclude}.

\begin{figure}
	\centering \includegraphics[width=9cm,height=5cm]{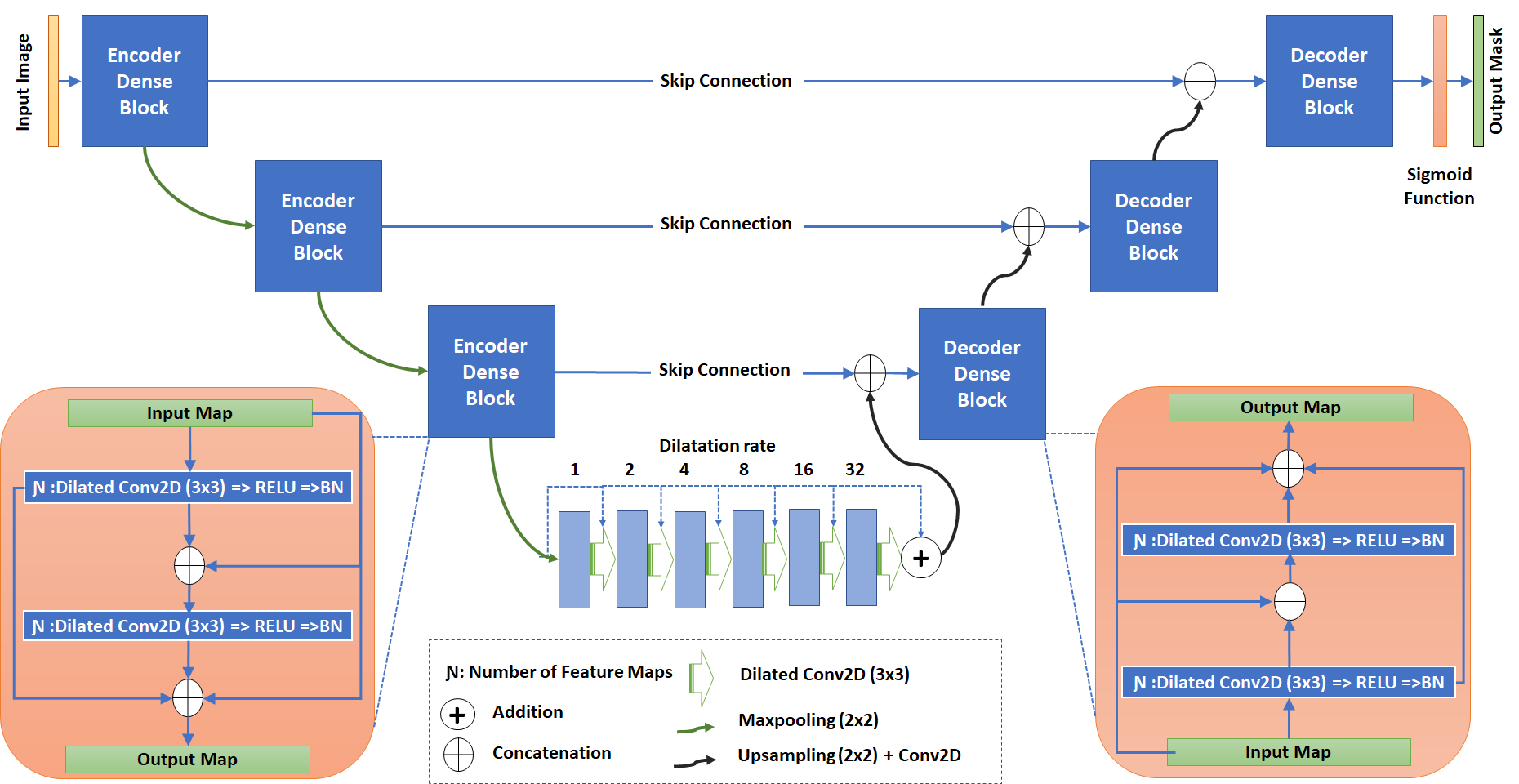}
	\caption{SkinNet architecture: in the bottleneck layer feature maps are convolved with a dilatation rate of 1-32. The dense blocks concatenate the feature maps from the previous layer with the current output feature map.}
	
	\label{fig:skinnet}
\end{figure}

\section{Methods}
\label{sec:methods}
Incorporating both local and global information, is generally beneficial for any segmentation task. However, in a conventional U-Net, the lowest level of the network has a small receptive field which prevents the network from extracting features that capture non-local image information. Dilated convolutions \cite{Dilated} provide a suitable solution to this problem. They introduce an additional parameter, called the dilation rate, to convolution layers, which defines the spacing between values in a kernel. This helps dilate the kernel such that a 3x3 kernel with a dilation rate of 2 will have a receptive field size equal to that of a 5x5 kernel. Additionally, this is achieved without any increase in complexity, as the number of parameters associated with the kernel remains the same which prevents the network from extracting features that capture non-local image information. Furthermore, multi-scale image information in the network, we replaced normal convolution layers at each level of both the encoder and decoder-branches, with densely connected convolution layers. In dense convolution blocks, the the inputs to each layer is a concatenation of outputs (or feature maps) from all preceding convolution layers. The overall architecture of SkinNet is summarized in Fig.2. The dense blocks show in blue boxes comprise two convolution layers with a kernel size of 3x3. At the bottleneck, encoded features are convolved using different dilation rates. The input images are normalized and resized to $512\times512$ pixels. We also performed data augmentation in order to boost the performance of SkinNet. The input images are randomly augmented using various image transformation techniques such as rotation, flipping, color shifting, translation and scaling operations.  

\subsection{Loss Function}
To measure the performance of the model, we defined a dice coefficient loss function which is summed over the classes:
\begin{equation}
\zeta(y, \hat{y})  = 1- \sum_{k}\frac{\sum_{n}y_{nk} \hat{y}_{nk}}{\sum_{n}y_{nk} + \sum_{n}\hat{y}_{nk}}
\end{equation}
$\hat{y}_{nk}$ denotes the output of the model, where $n$ runs over all pixels and $k$ runs over the classes (in our case, background vs. skin lesion). The ground truth masks are one-hot encoded and denoted by $y_{nk}$. We take one minus the dice coefficient in order to constrain the loss toward zero. 

% \subsection{Dataset}
% In order to evaluate the performance of SkinNet, we trained it on the ISBI 2017 challenge dataset \cite{ISBI}, which includes 2000 dermoscopic images and the corresponding lesion masks. The images in the dataset are of various dimensions form $1022\times767$ to $6688\times4439$. The lesion types involved include nevus, seborrhoeic keratosis, and malignant melanoma. In addition to the training set, the organizers also provided a validation dataset that includes 150 images and an additional test dataset with 600 images for final evaluation.

\subsection{Evaluation and Results}

In order to evaluate the performance of SkinNet, we trained it on the ISBI 2017 challenge dataset \cite{ISBI}, which includes 2000 dermoscopic images and the corresponding lesion masks. The organizers also provided a validation dataset that includes 150 images and an additional test dataset with 600 images for final evaluation. Lesion segmentation accuracy of our network was evaluated with respect to provided ground truth masks, using two well known similarity metrics: the Dice similarity coefficient (DC) and the Jaccard index (JI). We also calculated the sensitivity (SE), specificity (SP) and Accuracy (AC) metrics \cite{Ma} in order to directly compare our method with the state-of-the-art, presented in recent studies. We trained the network using the Adam optimizer \cite{adam} with a batch size of 8. ADAM was chosen as it provides an elegant mechanism for adaptively changing the learning rate, based on the first and second-order moments of the gradient, at each iteration. The initial learning rate is set as 0.0001 and reduced during the training. The network trained with 5 folds cross validation on the training dataset and tested and validated with another 600 and 150 images. Fig.2. shows the average of training plot for 100 epochs and as it can be seen, there is no overfitting during the training process.  The results for the test dataset are also shown in Table 1. respectively.

% The metrics are defined as:
% which measures the overlap between two segmentation masks and is sensitive to the lesion size,
%, which denotes the average distance between two masks
% \begin{equation}
% DC (A,B) = \frac{2|A \cap B|}{|A|+|B|}
% \end{equation}

% \begin{equation}
% JA (A,B) = \frac{|A \cup B|}{|A \cap B|}
% \end{equation}

% Fig.2. shows the average of training plot for 100 epochs and as it can be seen, there is no overfitting during the training process.

% \begin{figure}
% 	\centering \includegraphics[width=8.5cm,height=5.5cm]{SkinNet_graph}
% 	\caption{Training loss, dice and Jaccard curves for 5 fold cross validation (average).}
	
% 	\label{fig:skinnet}
% \end{figure}

\begin{figure}
    \centering
    \subfloat{{\includegraphics[width=4.3cm, height=4cm]{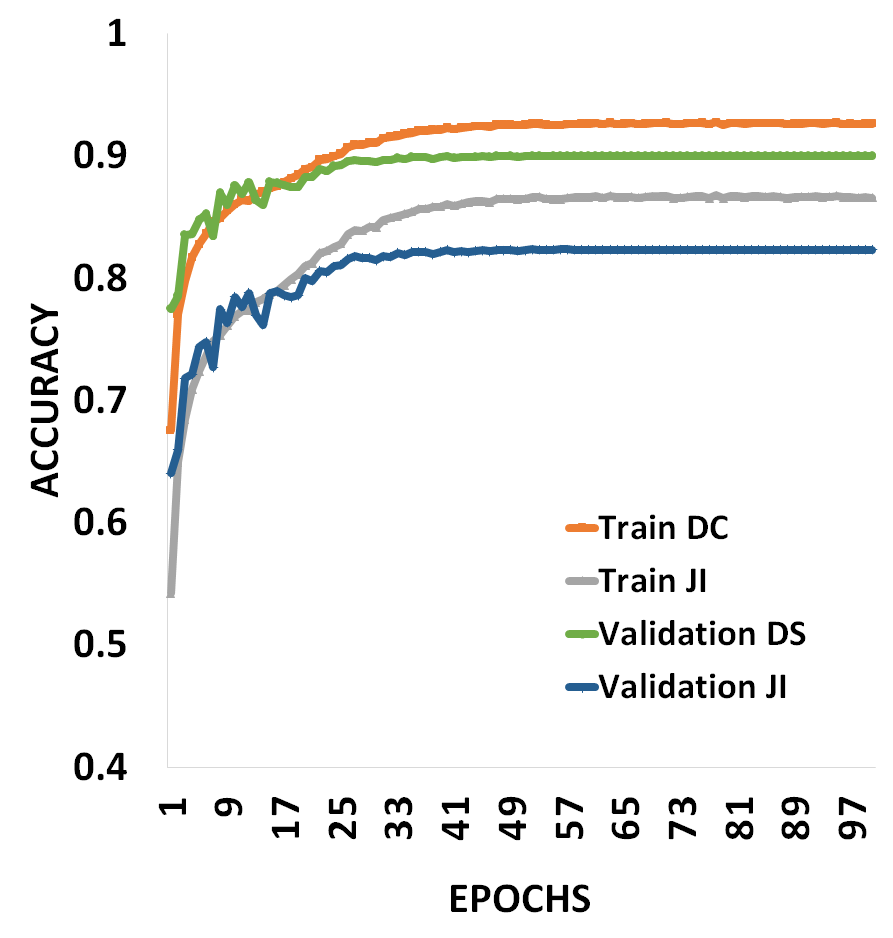} }}
    \subfloat{{\includegraphics[width=4.3cm, height=4cm]{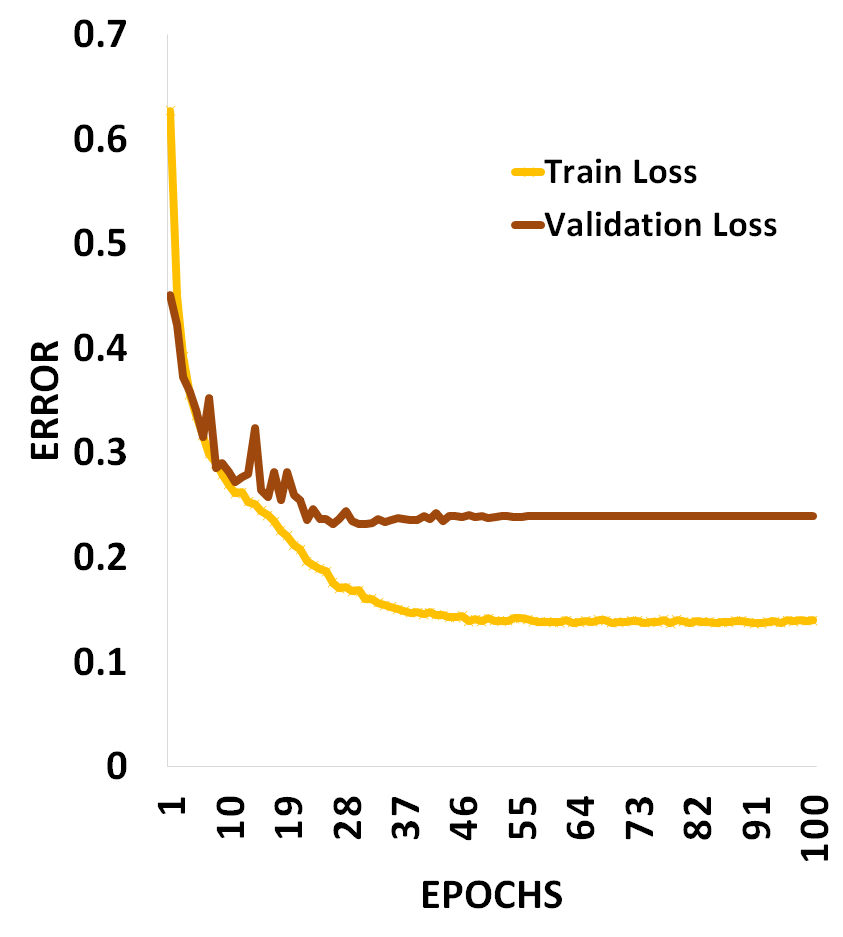} }}
    \caption{Training and validation loss, DS and JI curves for 5 fold cross validation (average).}
    \label{fig:example}
\end{figure}

\begin{table}[]
\small
	\centering
	\caption{Lesion segmentation performances of different frameworks.}
	\label{my-label}
	\begin{tabular}{|l|p{0.8cm}|p{0.8cm}|p{0.8cm}|p{0.8cm}|p{0.8cm}|}
		\hline
		\textbf{Method}       & \textbf{AC} & \textbf{DC}    & \textbf{JI}    & \textbf{SE}    & \textbf{SP} \\ \hline \\[-1em]
		Yading Yuan \cite{Yuan}  & 0.934       & 0.849          & 0.765          & 0.825          & 0.975       \\   \hline
		Math Berseth & 0.932       & 0.847          & 0.762          & 0.820          & 0.978       \\   \hline
		Auto-ED \cite{Attia}      & 0.936       & 0.824          & 0.738          & 0.836          & 0.966       \\   \hline
		LIN \cite{s18020556}          & 0.952        & 0.839          & 0.753          & 0.855          & 0.974      \\   \hline
		SkinNet      & 0.932       & \textbf{0.851} & \textbf{0.767} & \textbf{0.930} & 0.905       \\ \hline
	\end{tabular}
\end{table}

\section{Discussion and Conclusion}
\label{sec:conclude}
While various network configurations have been proposed for skin lesion segmentation, to the best of our knowledge none of them employ an architecture identical to SkinNet. In this paper, we proposed a novel deep learning approach which exploits both local and global image information, for skin lesion segmentation. Our method achieved higher JI and DC compared to other state-of-the-art methods (as shown in Table 1) which used various types of CNN architectures. The dense blocks incorporate multiscale level information effectively in the encoder and decoder branches of the network. The JI and DC values achieved by our network were 76.67 and 85.10\%, respectively. Additionally, SkinNet achieved the highest sensitivity (93\%), outperforming the rest.

\section*{Acknowledgment}
The authors gratefully acknowledge the support of Emerging Field Intuitive (EFI) project of Friedrich Alexander University Erlangen-Nuremberg.

% \end{thebibliography}

\bibliographystyle{IEEEtran}

\bibliography{0000}

\end{document}